\documentclass[fleqn,10pt]{wlscirep}
\usepackage[utf8]{inputenc}
\usepackage[T1]{fontenc}
\usepackage{graphicx} 
\usepackage{caption} 
\usepackage{subcaption} 
\usepackage{setspace} 
\usepackage{xcolor}

\title{Representational Ethical Model Calibration}

\author[1*]{Robert Carruthers}
\author[2] {Isabel Straw}
\author[2] {James K Ruffle}
\author[3] {Daniel Herron}
\author[2] {Amy Nelson}
\author[4] {Danilo Bzdok}
\author[1] {Delmiro Fernandez-Reyes}
\author[2] {Geraint Rees}
\author[2*] {Parashkev Nachev}

\affil[1]{Department of Computer Science, University College London, London, UK}
\affil[2]{UCL Queen Square Institute of Neurology, University College London, London, UK}
\affil[3]{Research and Development, NIHR University College London Hospitals Biomedical Research Centre, London, UK. }
\affil[4]{Department of Biomedical Engineering, Faculty of Medicine, McGill University, Montreal, Canada. }

\affil[*]{robert.carruthers.20@ucl.ac.uk or p.nachev@ucl.ac.uk}

\begin{abstract}
Equity is widely held to be fundamental to the ethics of healthcare. In the context of clinical decision-making, it rests on the comparative fidelity of the intelligence -- evidence-based or intuitive -- guiding the management of each individual patient. Though brought to recent attention by the individuating power of contemporary machine learning, such epistemic equity arises in the context of any decision guidance, whether traditional or innovative. Yet no general framework for its quantification, let alone assurance, currently exists. Here we formulate epistemic equity in terms of model fidelity evaluated over learnt multi-dimensional representations of identity crafted to maximise the captured diversity of the population, introducing a comprehensive framework for \textit{Representational Ethical Model Calibration}. We demonstrate use of the framework on large-scale multimodal data from UK Biobank to derive diverse representations of the population, quantify model performance, and institute responsive remediation. We offer our approach as a principled solution to quantifying and assuring epistemic equity in healthcare, with applications across the research, clinical, and regulatory domains.
\end{abstract}
\begin{document}

\flushbottom
\maketitle

\thispagestyle{empty}


\section*{Introduction}
Medicine has always been personal, concerned with the individual patient whose specific complaint the physician is asked to address. Under pressure to render the underlying intelligence explicit, objective, replicable, and cumulative, evidence-based medicine has shifted the focus to large populations, guiding clinical management by the parameters of simple statistical models that discard individual variation as noise \cite{sackett1997evidence,greenhalgh2014evidence}. The resultant gain in population-level fidelity may or may not be associated with a loss at the individual level: we do not know, because the studies that generate care policies and those that evaluate them adopt the same inferential approach. 

\textcolor{black}{This blind spot extends to \textit{equity} of care: the universal obligation, rooted in Aristotle's notion of \textit{epieikeia} \cite{crisp2014aristotle}, to seek the best achievable outcome for each individual patient. The care recommended by a model is inequitable as far as it fails to fulfil a patient's individual potential for recovery and health\cite{WHO}. In neglecting individuality, conventional evidence-based medicine conceals our success or failure in maintaining such equity. This defect is all the more important for being \textit{epistemic}, of the upstream knowledge from which all downstream clinical action derives.}

\textcolor{black}{Until the advent of machine learning, this major ethical problem had no obvious remedy. But it is now clear that richly expressive models of high-dimensional data can characterize populations with greater fidelity to the individual \cite{xiao2018opportunities, vasiljevic2021smoking,bica2021real,bzdok2018machine}. Whether implicitly or explicitly, such models describe patients in terms of more closely individuating \textit{subpopulations} identified by multiple interacting characteristics, whose distinct structure may be directly material to clinical care, interfere with our ability to determine its optimal form, or both. By revealing differences between subpopulations, machine learning casts a brighter light on epistemic equity than crude population descriptions could provide, and enables us to pursue our deep moral obligation to assure it.} 

Though a matter of intense study in other domains, there is no accepted framework for defining, diagnosing, or quantifying epistemic equity arising in the guidance of clinical care by algorithmic models, whether simple or complex, traditional or novel, and neither regulatory nor professional bodies currently provide for it. Here we propose such a framework, termed Representational Ethical Model Calibration.

\textcolor{black}{Any quantitative framework here must operationalise the notions of the epistemic equity of a model and the descriptive identity of a patient. We define the former as equal maximisation of model fidelity across the population: where the available knowledge is plausibly invariant, equity means equality; where it varies under some external constraint outside our power to address, equity means equal departure from the attainable maximum. We define the latter as any set of replicable distinguishing characteristics material to the specific healthcare context. For example, the equity of a classifier for detecting ischaemic injury on a brain scan might be measured by balanced accuracy evaluated as a function of age. Observing systematic variation of accuracy with age raises the possibility of inequitable performance across patients so identified.} 

\textcolor{black}{Many quantitative indices of model fidelity and its variation exist: the optimal choice will vary with the specific application. The appropriate criteria of identity, however, are not so easy to determine. The use of simple descriptors such as age, sex, and ethnicity, taken in isolation, presupposes that they are sufficiently individuating. But, as we have seen, a patient will typically belong to a distinct, replicable -- and therefore learnable -- subpopulation defined by the interaction of multiple characteristics. The underperformance of a model in such a subpopulation may not be evident from examination of single characteristics alone. Any principled notion of equity obviously cannot exclude groups whose defining identity eludes simple description. Indeed, there is increasing evidence that it is precisely those falling in the intersectional \textit{faultlines} between traditionally recognized groups that may be most vulnerable \cite{lau1998demographic,dibenigno2014beyond,li2005factional,thatcher2011demographic, bambra2022placing}. Moreover, neither the total number nor the nature of the relevant identifying characteristics may be limited \textit{a priori}. If a social, environmental, demographic, physiological, pathological or any other replicable distinguishing characteristic -- whether self-assigned or externally measured -- has a systematic impact on clinical outcomes, we have a moral duty to examine it. Naturally, the wide descriptive space so defined may not be easily navigable. But we can employ \textit{representation learning} \cite{bengio2013representation} to derive rich yet succinct descriptions of the population that render its diversity surveyable.}

\textcolor{black}{Our proposed framework therefore combines the evaluation of model performance against identifying descriptors -- ethical model calibration -- with the derivation of descriptors through representation learning that optimally capture the diversity of the population. In relating observed to ideal performance, it is kin with statistical model calibration \cite{niculescu2005predicting}. It enables the epistemic equity of a model to be judged against identities defined as richly and comprehensively as available data allow.}

\begin{figure}[]
\centering
\includegraphics[width=12cm]{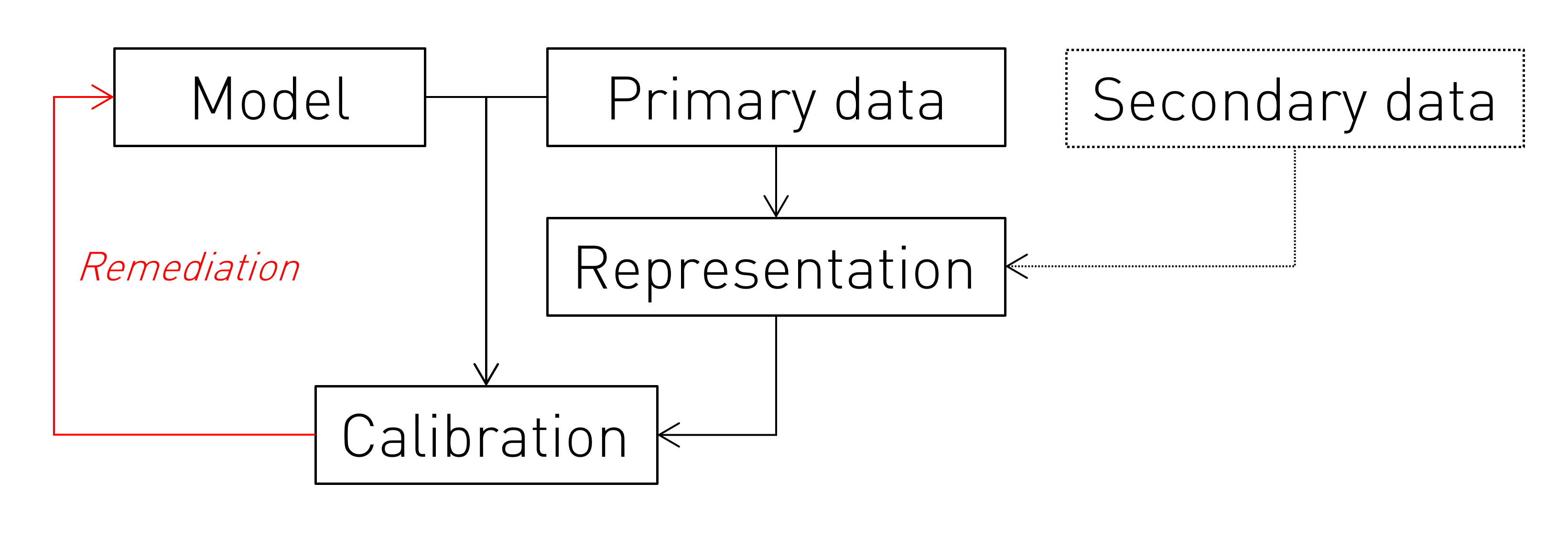}
\caption{\textbf{The Representational Ethical Model Calibration Framework.} The fidelity of a candidate model with respect to subpopulations identified by representation learning (performed on either primary or secondary data) is quantified in an ethical calibration step that informs appropriate remedial action, within an iterative process repeated until an agreed criterion of model equity is reached.}
\label{fig:emc}
\end{figure}

\textcolor{black}{Application of the framework is illustrated in Figure \ref{fig:emc}. In brief, the fidelity of a given model, quantified by the metric most suited to its task, is evaluated against a succinct description of the population derived from learnt representations of the same (primary) data or other (secondary) data drawn from the same domain. Systematic differences in performance across the population identify potential inequities, and trigger remediation -- action to correct the disparity or limit its downstream impact -- by any applicable mechanism such as acquiring more data, modifying the model, or limiting its application. The cycle of calibration and remediation may be repeated until a result satisfactory on some agreed criterion is obtained. Extending the foregoing example, an ischaemic stroke classifier may be found to perform poorly within a distinct subpopulation with a characteristic age interval and spectrum of co-morbidities. Identifying this subpopulation directs action on the data, the model, or the scope of application, until calibration shows equity has been achieved. Where the representations are based on a generative model of the primary data, it enables immediate remediation by augmenting model retraining with synthetic data from the under-performing subpopulation.}

This approach is applicable to any model, whether conventional or machine learning-based, any metric of performance and its disparity, and any method of representation learning \cite{weng2019representation,girkar2018predicting,landi2020deep,miotto2016deep}. It leaves the nature of the remediation open, to be chosen as specific circumstances dictate, and distinguishes remediation from the calibration used to guide it. Here we demonstrate its use, end-to-end, in the context of predicting glycaemic control -- as indexed by glycated haemoglobin (HbA1c) concentrations \cite{world2011use} -- from large-scale, high-dimensional data in UK Biobank \cite{sudlow2015uk}. We choose glycaemic control owing to the importance of glucose intolerance and complex patterns of its susceptibility. We employ deep representation learning based on autoencoders owing to their architectural simplicity and expressivity, and established applications in healthcare \cite{xiao2018opportunities}. 

\textcolor{black}{We show how the framework can be used to detect the systematic epistemic inequity of a model with respect to subpopulations concealed by the richness of their identity, and guide remediation in pursuit of more equitable model performance. Although model epistemic equity is only one aspect of equity, itself only one aspect of medical ethics, the position of models at the apex of evidence-based clinical decision-making lends the highest ethical significance to the equity of their performance. Our results are relevant to the domains of quantitative ethics, multidimensional fairness, and the regulation of mathematical models in healthcare \cite{challen2019artificial}.}

\section*{Results}

\subsection*{Associations of impaired glycaemic control}
\begin{figure}
    \centering
    \includegraphics[width=16cm]{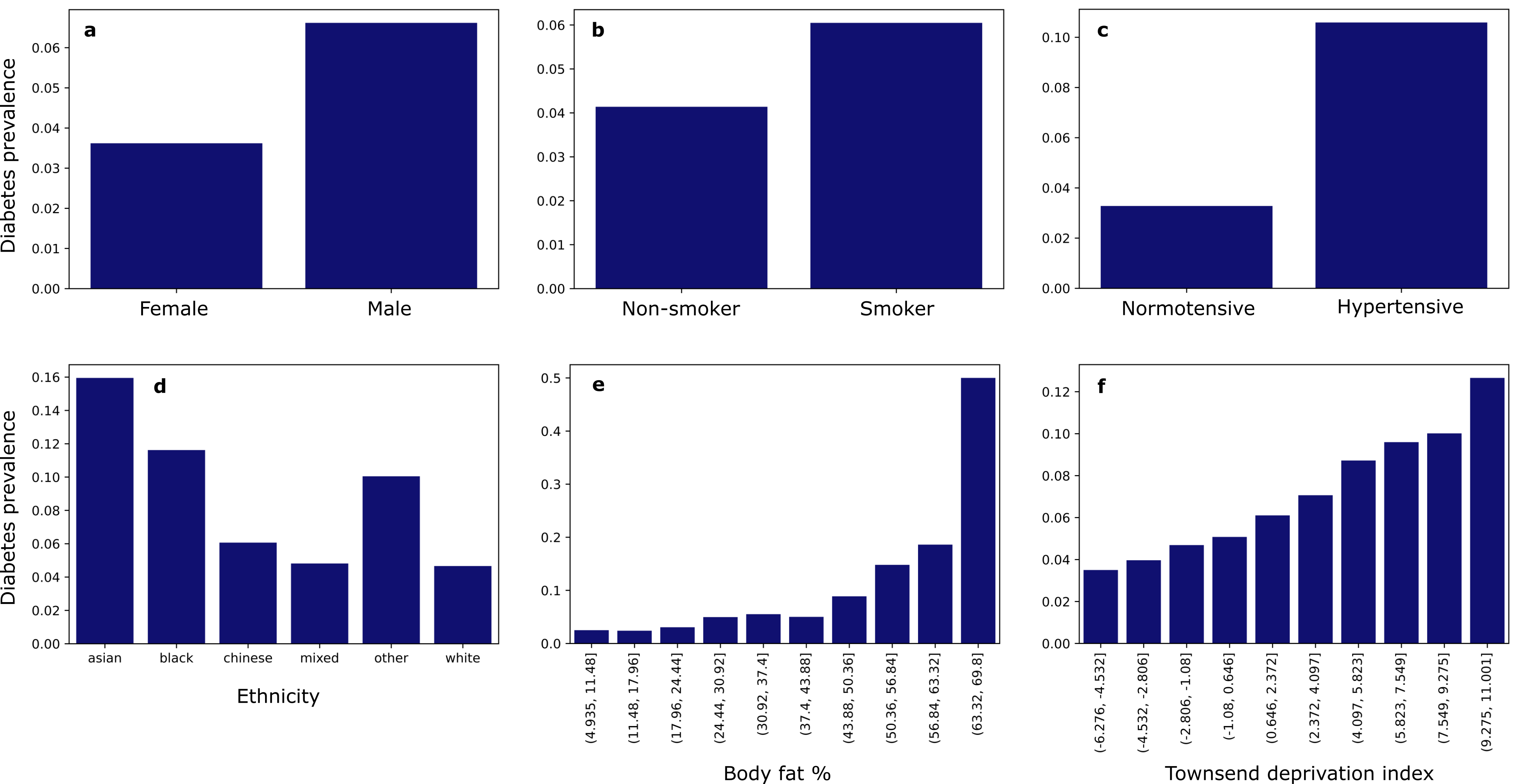}
    \caption{\textbf{Diabetes prevalence by variable.} Higher prevalence was seen in males (a), smokers (b), those with high blood pressure (c), certain ethnicities (d), those with higher body fat \% (e), and the more deprived (f). }
    \label{fig:diab_prev}
\end{figure}

\begin{figure}
    \centering
    \includegraphics[width=10cm]{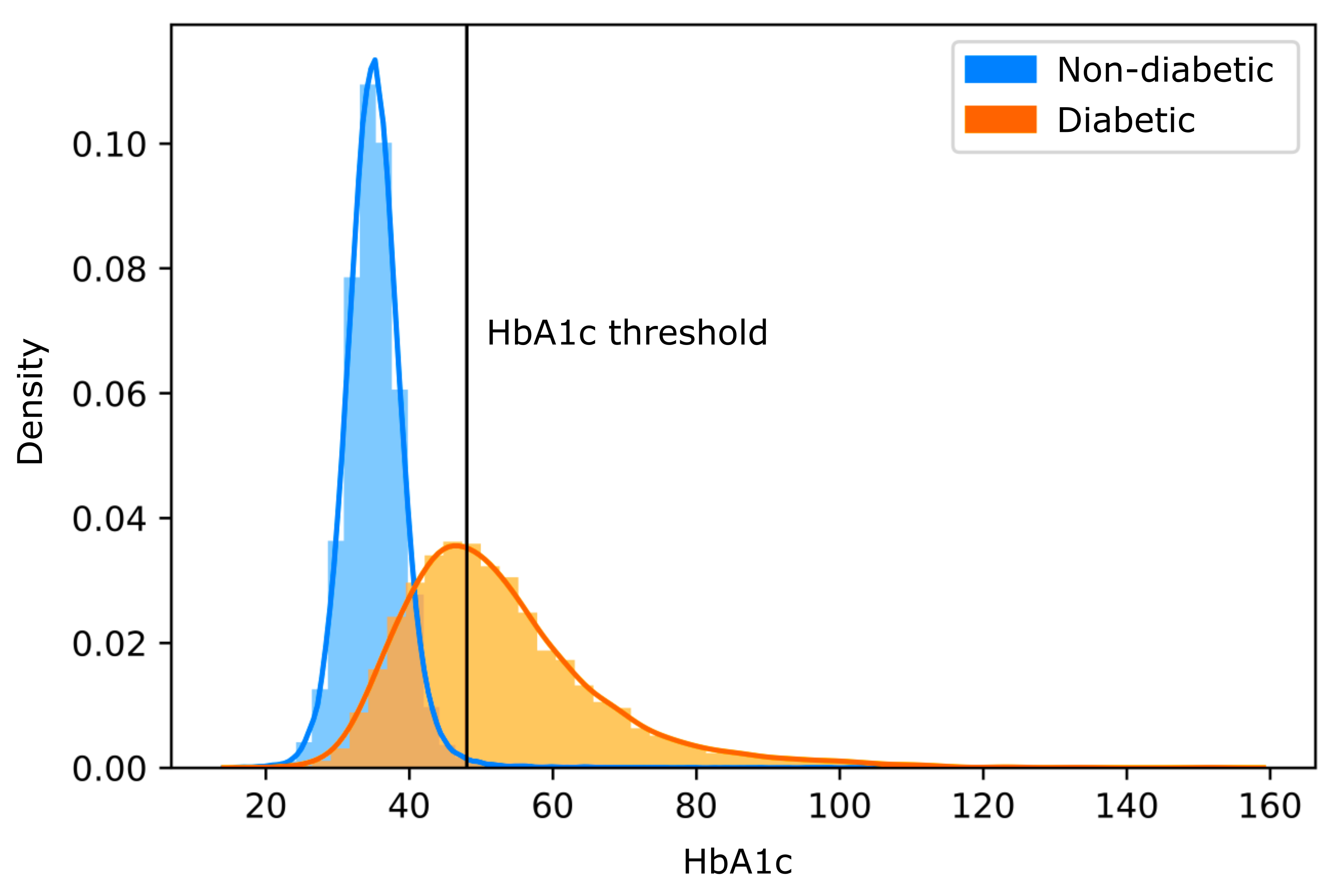}
    \caption{\textbf{Relationship of diabetes and glycated haemoglobin (HbA1c).} Those without diabetes tended to have HbA1c below the diagnosis threshold of 48, while those with diabetes had a wide range of HbA1c both above and below the threshold.}
    \label{fig:diab_hba1c}
\end{figure}

We evaluated a random selection of UK Biobank records split into 150,000 training and 50,000 validation sets, including a range of demographic, social, lifestyle, physiological, and morbidity features potentially relevant to glycaemic control (see Methods). The commonest cause of impaired glycaemic control -- diabetes -- showed variation with sex, smoking, hypertension, ethnicity, body fat composition, and social deprivation consistent with previous data from populations with a similar age distribution (Figure \ref{fig:diab_prev}). A diagnosis of diabetes was associated with higher and more widely dispersed HbA1c, reflecting variable success in the clinical management of the underlying disorder (Figure \ref{fig:diab_hba1c}). These observations suggest predictive models of HbA1c based on this data can be considered representative of a plausible real-world modelling scenario.

\subsection*{Basic ethical model calibration}

A regression model based on a conventional fully-connected feed-forward net with three hidden layers, an architecture chosen for its controllable flexibility (see Methods), was evaluated across the population as a whole. Near-identical root mean squared errors were observed on the training and validation sets -- 6.099 and 6.097, respectively -- corresponding to normalised root mean squared errors (NRMSE) of 0.169 for both. Nonetheless, examination of performance independently stratified by sex, smoking, and deprivation revealed substantial disparities in model fidelity (Table \ref{tab:diab_hba1c_bias}), with evident underperformance for men, smokers, and the socially more deprived.

\subsection*{Representational ethical model calibration}

\begin{figure}
    \centering
    \includegraphics[width=16cm]{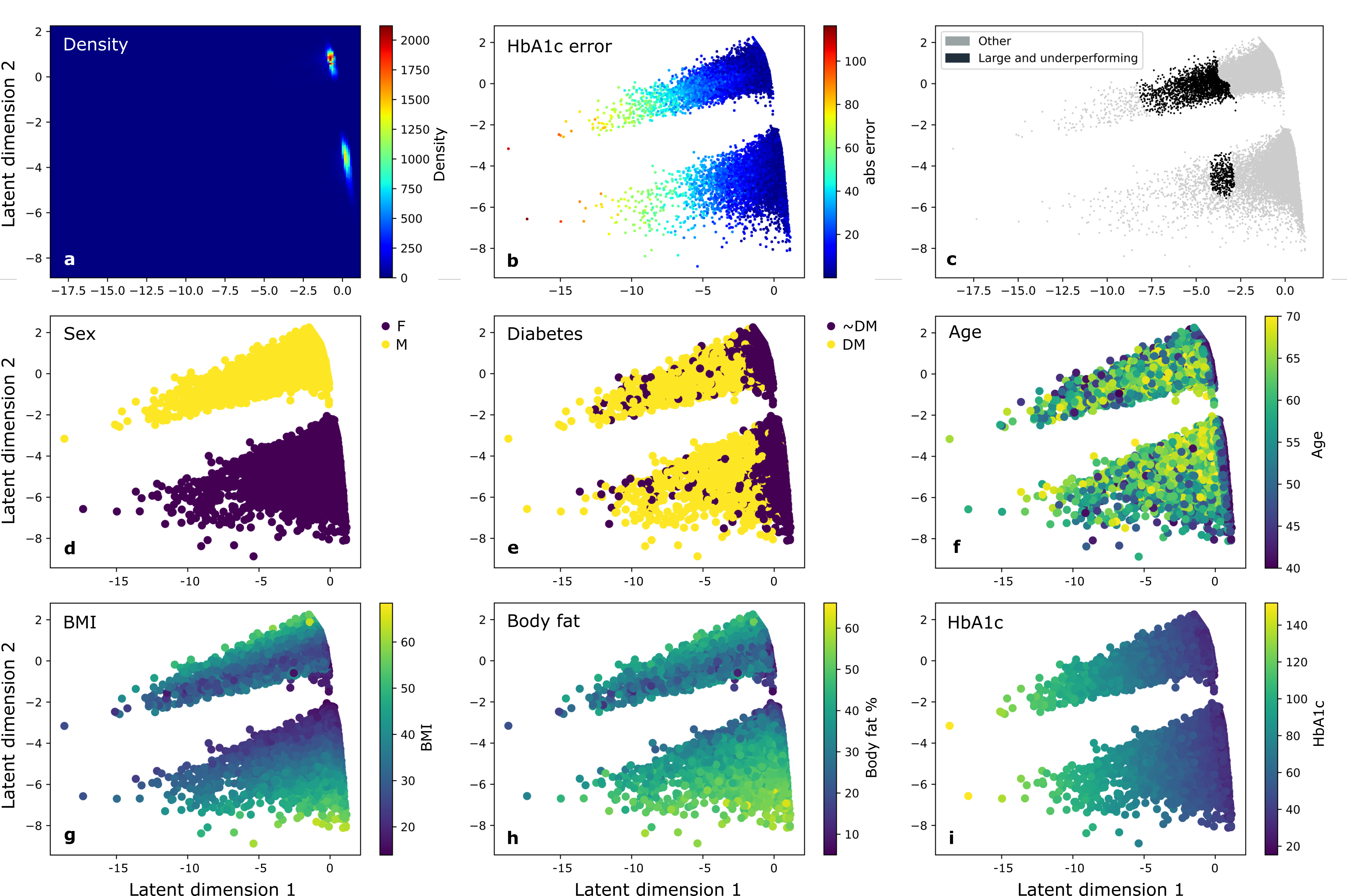}
    \caption{\textbf{Two-dimensional latent space.} The space is coloured by data density (a), model error (b,c), and the values of selected variables (d-i). The space appears to be clustered by sex (d). The largest groups in the worst 25th performance percentile are shown in (c), associated with higher levels of HbA1c (i).}
    \label{fig:diab_hba1c_latent}
\end{figure}

To permit the identification of underperformance localised to more complex subpopulations defined by the interactions of multiple factors, we used an autoencoder to embed participants in a two-dimensional latent representational space that compactly described their high-dimensional similarities and differences (see Methods). Labelling the embedding by individual regression errors (Figure \ref{fig:diab_hba1c_latent}b) revealed potentially structured variation in fidelity; labelling it by key descriptive features revealed its organisation (Figure \ref{fig:diab_hba1c_latent}d-i). To facilitate the identification of a tractable number of characteristic subpopulations, the latent space was segmented into fifty groups using a Gaussian Mixture Model (GMM). Examining the five largest groups in the bottom 25th regression performance centile revealed a diversity of patterns of individual features, 
most of them shown on permutation testing to be significantly different from the rest of the population (Table \ref{tab:diab_hba1c_permut}). Crucially, model fidelity varied more widely than across basic features of ethical concern, and reached lower values, illustrating the need for evaluating feature interactions in the quantification of equity.   

\subsection*{Remediation}

\begin{figure}
    \centering
    \includegraphics[width=17cm]{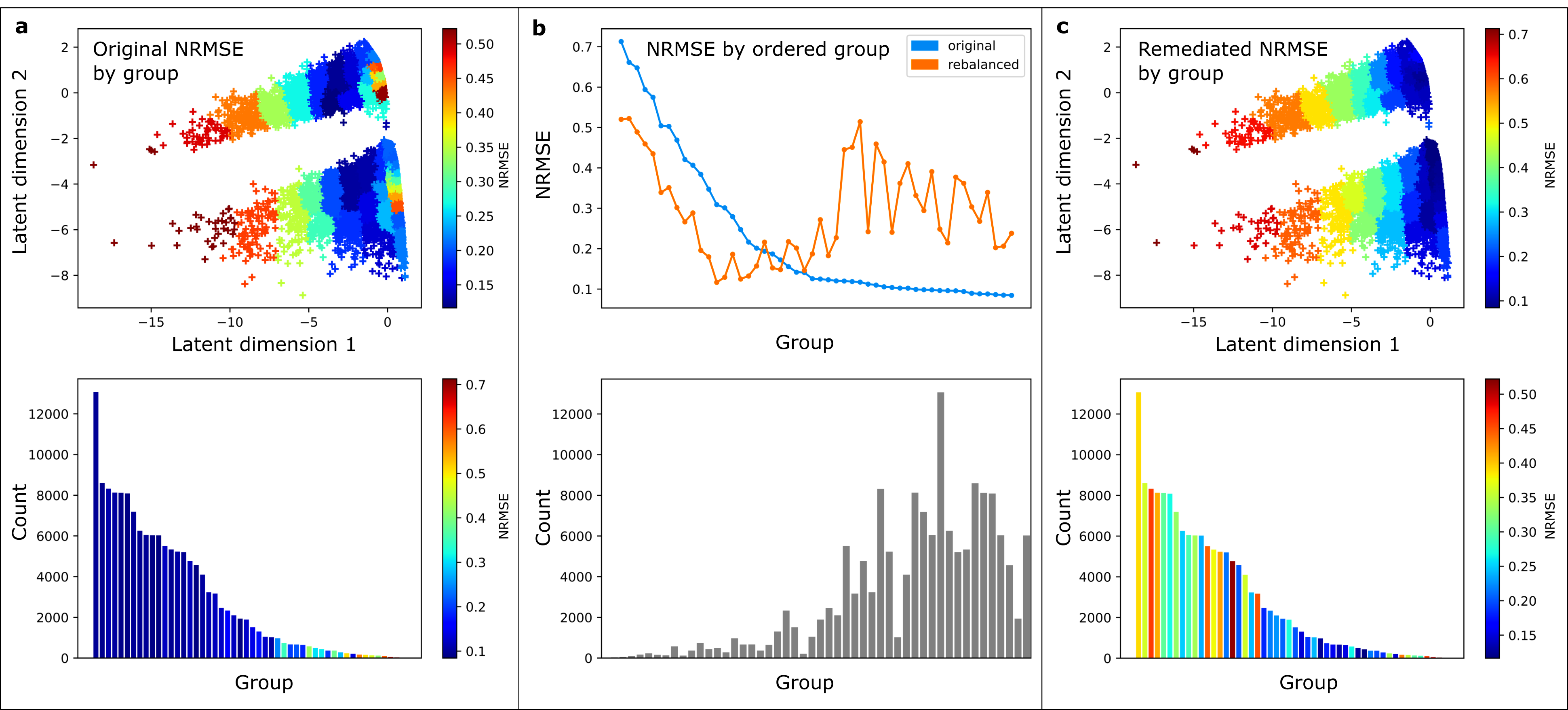}
    \caption{\textbf{Model performance by GMM group in the latent space.} Performance is shown before (a) and after (c) remediation. The top panel of (a) shows model performance by group, while the bottom panel shows the group counts. The model showed mostly even performance across groups. The top panel of (b) shows the effect of remediation. Lower-performing (higher NRMSE) groups show improvements, but the better-performing groups got significantly worse. The bottom panel of (b) shows group counts in descending order of the original NRMSE. It can be seen that performance decreases occurred in high-volume groups. The performance distribution worsened overall, shown in (c), and this would likely offset any gain in equity.}
    \label{fig:diab_hba1c_debiasing}
\end{figure}

\begin{figure}
    \centering
    \includegraphics[width=16cm]{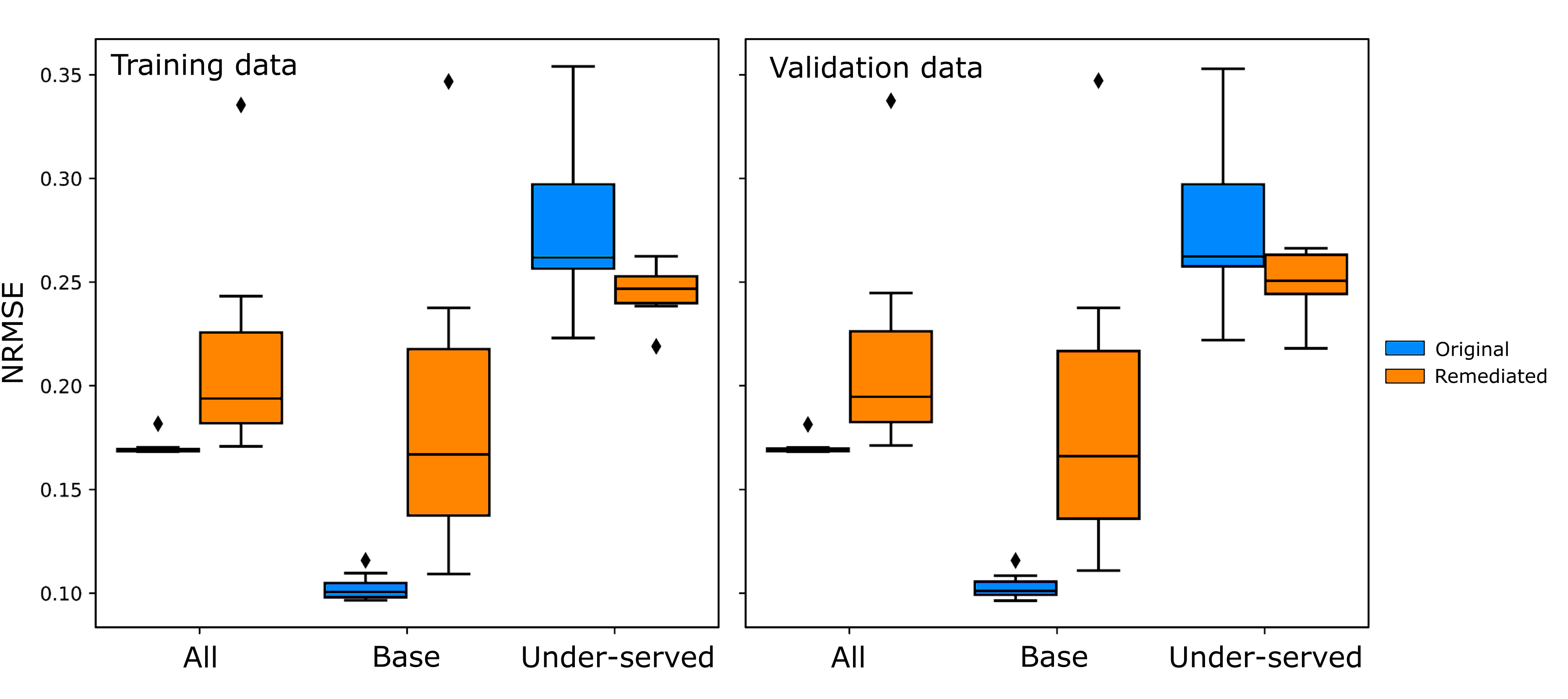}
    \caption{\textbf{Effect of remediation.} NRMSE is shown for the whole dataset, the base group and the under-served group, before and after remediation, over n=10 trials, on training and validation data. Performance was worse on the under-served group, and this improved after rebalancing. However, there was a high cost in base group performance. See differences in Table \ref{tab:diab_hba1c_results}.}
    \label{fig:diab_hba1c_results}
\end{figure}

\begin{figure}
    \centering
    \includegraphics[width=16cm]{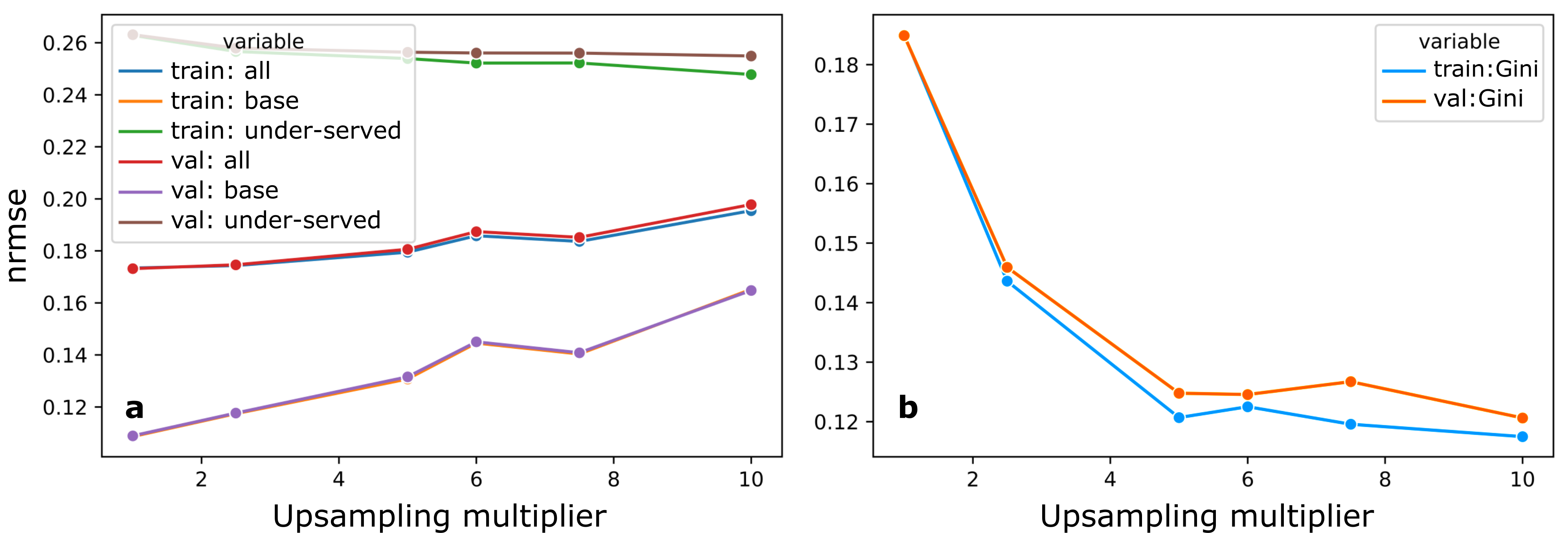}
    \caption{\textbf{Effect of upsampling multiplier on performance metrics.} Panel (a) shows NRMSE for the entire dataset, the base group and the under-served group, for training and validation sets. Panel (b) shows the Gini coefficient. In both training and validation sets, increasing the upsampling multiplier improved model performance on the under-served group, while negatively affecting performance on the base group and overall. The Gini coefficient tended to drop as upsampling increased, mostly indicating increased equity in the distribution for higher levels of upsampling.}
    \label{fig:diab_hba1c_oversamp}
\end{figure}

\textcolor{black}{Revealing the pattern of inequitable performance allows us to target our efforts at remediation. The optimal approach to remedying inequitable models -- in healthcare and elsewhere -- is the subject of intense study, and will vary circumstantially in feasibility and effectiveness. Here we illustrate only one approach to remediation, focused on model training. The subpopulations exhibiting higher than median NRMSE were designated as under-served. A simple strategy of remediation was then applied, oversampling these subpopulations in model training, generating a very different pattern of performance with less pronounced disparities across the population, but at the cost of reduced fidelity overall (Figure \ref{fig:diab_hba1c_debiasing} and \ref{fig:diab_hba1c_results}). These observations can be formalised in terms of NRMSE, and standard indices of distributional equality such as the Gini coefficient (Tables \ref{tab:diab_hba1c_results} and \ref{tab:diab_hba1c_gini}). Varying the degree of oversampling moved NRMSE and Gini coefficient scores in the expected directions, as shown
in Figure \ref{fig:diab_hba1c_oversamp}. Performance disparities persisted across groups, such as men and women, defined by dimensions other than those selected for remediation. Note our objective here is not to devise or implement an optimal approach to remediation, but to show how the calibration and remediation processes relate. An optimal approach would improve equity without deleterious impact on other groups or overall performance.}

\begin{table}[]
\small
\singlespacing
    \centering
    \begin{tabular}{ccccccc}
    \cline{2-7}
         \multicolumn{1}{c|}{} & \multicolumn{2}{c|}{Sex} & \multicolumn{2}{c|}{Smoking} & \multicolumn{2}{c|}{Deprivation}\\ \cline{2-7}
         \multicolumn{1}{c|}{} & \multicolumn{1}{c}{F} & \multicolumn{1}{c|}{M} & \multicolumn{1}{c}{No} & \multicolumn{1}{c|}{Yes} & \multicolumn{1}{c}{Low} & \multicolumn{1}{c|}{High}\\ \cline{1-7}
         \multicolumn{1}{|c|}{Training} & 0.152 &  \multicolumn{1}{c|}{0.186} & 0.162 & \multicolumn{1}{c|}{0.176} & 0.156 & \multicolumn{1}{c|}{0.187}\\
         \multicolumn{1}{|c|}{Validation} & 0.151 & \multicolumn{1}{c|}{0.187} & 0.161 & \multicolumn{1}{c|}{0.177} & 0.158 & \multicolumn{1}{c|}{0.185}\\ \cline{1-7}
    \end{tabular}
    \caption{\textbf{Stratified model performance to demonstrate inequity.} NRMSE is shown across three example variables. The deprivation index was binarized into high and low by mean split. The model's overall training NRMSE was 0.169. Consistent performance differences are shown, with the model performing better (lower NRMSE) for women, non-smokers and the less deprived.}
    \label{tab:diab_hba1c_bias}
\end{table}

\begin{table}[]
\small
\singlespacing
    \centering
    \begin{tabular}{|c|c|c|c|c|c|}
    \cline{2-6}
         \multicolumn{1}{c|}{}  & \multicolumn{5}{c|}{Group} \\ \cline{2-6}
         \multicolumn{1}{c|}{} & 22 & 2 & 41 & 33 & 23\\ \cline{1-6}
         High Blood Pressure & 0.0010 & 0.0010 & 0.0010 & 0.0010 & 0.0010\\
         Smoking & 0.0010 & 0.0010 & 0.0010 & 0.1029 & 0.0010\\
         BMI & 0.0010 & 0.0010 & 0.0010 & 0.0010 & 0.0010\\
         HbA1c & 0.0010 & 0.0010 & 0.0010 & 0.0010 & 0.0010\\
         Age & 0.0010 & 0.0010 & 0.0010 & 0.0010 & 0.0010\\
         Ethnicity: white & 0.0579 & 0.0010 & 0.0010 & 0.0010 & 0.0010\\
         Townsend Deprivation Index & 0.1349 & 0.0010 & 0.0010 & 0.0040 & 0.0010\\ \cline{1-6}
    \end{tabular}
    \caption{\textbf{Permutation test results.} Selected permutation test results (p-values) are shown for the five largest groups in the worst 25th performance percentile for prediction. P-values below a given significance level indicate that the group's mean was significantly different from the remainder of the data for that variable. Applying the Benjamini-Hochberg procedure to control the False Discovery Rate across multiple tests resulted in $85$ significant results out of $110$ tests at $\alpha=0.05$. }
    \label{tab:diab_hba1c_permut}
\end{table}

\begin{table}[]
\small
\singlespacing
    \centering
    \begin{tabular}{c|c|c||c|c|}
    \cline{2-5}
         &  \multicolumn{2}{c||}{Mean Difference in NRMSE} & \multicolumn{2}{c|}{Standard Deviation} \\ \cline{2-5}
         & Training & Validation & Training & Validation\\ \cline{1-5}
         \multicolumn{1}{|c|}{All} & -0.0411 & -0.0420 & 0.0466 & 0.0470\\
         \multicolumn{1}{|c|}{Base} & -0.0817 & -0.0809 & 0.0666 & 0.0667\\
         \multicolumn{1}{|c|}{Under-served} & 0.0315 & 0.0268 & 0.0335 & 0.0300\\ \cline{1-5}
    \end{tabular}
    \caption{\textbf{Difference in Normalised Root Mean Square Error (NRMSE) before and after rebalancing.} Mean differences and standard deviations of differences are presented across n=10 trials. Results are presented for the entire dataset, the base group and the under-served group. A positive difference indicates better model performance. It can be seen that base group performance worsened on average while under-served group performance improved. NRMSE values are illustrated in Figure \ref{fig:diab_hba1c_results}.}
    \label{tab:diab_hba1c_results}
\end{table}

\begin{table}[]
\small
\singlespacing
    \centering
    \begin{tabular}{c|c|c||c|c|}
    \cline{2-5}
         &  \multicolumn{2}{c||}{Mean Gini Coefficient} & \multicolumn{2}{c|}{Standard Deviation} \\ \cline{2-5}
         & Training & Validation & Training & Validation\\ \cline{1-5}
         \multicolumn{1}{|c|}{Original} & 0.1594 & 0.1595 & 0.0163 & 0.0170\\
         \multicolumn{1}{|c|}{Rebalanced} & 0.1466 & 0.1470 & 0.0309 & 0.0299\\
         \multicolumn{1}{|c|}{Difference} & 0.0128 & 0.0124 & 0.0321 & 0.0324\\ \cline{1-5}
    \end{tabular}
    \caption{\textbf{Gini coefficients.} Mean Gini coefficients, differences and standard deviations are presented, before and after rebalancing, across n=10 trials. A Gini coefficient of zero indicates equality and a Gini coefficient of one indicates maximum inequality. A positive difference can be be observed, which indicates a decrease in Gini coefficient on average and hence a more equitable distribution of performance across groups. However, the standard deviations indicate that this behaviour was not consistent.}
    \label{tab:diab_hba1c_gini}
\end{table}

\section*{Discussion}

\textcolor{black}{We have formulated a framework, Representational Ethical Model Calibration, for detecting and quantifying inequity in model performance distributed across subpopulations defined by multiple interacting characteristics. Central to the framework is the use of representation learning to compress and render navigable the high-dimensional space of patient diversity over which equitable performance must be evaluated. We have demonstrated ethical model calibration on large-scale UK-biobank data in the context of a common morbidity -- impaired glycaemic control -- and a simple, purely illustrative approach to remediation. Here we examine ten aspects of the conceptualisation, implementation, and application of our approach.}

\textcolor{black}{First, if epistemic equity cannot be assumed where, as here, the modelling task is comparatively simple and the data are balanced and abundant, the case for evaluating it explicitly across all models employed in healthcare ought to be hard to resist. The narrow recruitment mechanism makes a test based on UK Biobank conservative, for the underlying heterogeneity is likely to be less pronounced than that observed in clinical reality. Equally, our analysis shows that calibration across familiar, observed, unitary features such as demographics cannot reveal inequity across unfamiliar, latent, composite features accessible only through representation learning. A commitment to ethical model calibration implies a commitment to the representational kind.}

\textcolor{black}{Second, the epistemic equity of the models used to guide clinical care -- our concern here -- is obviously not the only kind of equity healthcare must consider. Unwarranted variation in clinical outcomes may arise from a wide diversity of procedural, cultural, social, economic, political, and regulatory factors that operate outside the realm of evidence-guided practice and need to be addressed independently from it \cite{challen2019artificial,vayena2018machine,yu2020one,singh2018deep,de2018clinically,xiao2018opportunities,ferryman2018fairness,weng2019representation,blank2017comparative,starke2021towards,char2018implementing,goldacre2014bad,vayena2018machine,mccradden2020ethical}. In focusing on epistemic equity we are not denying the importance of other kinds. But since action proceeds from belief, and belief in medicine strives to be objectively evidential, detecting and quantifying epistemic equity will always be a fundamental concern. Equally, in using the qualifier "ethical", we do not mean to imply that detecting equity, still less epistemic equity, exhausts the ethical realm, only to indicate the \textit{purpose} of the calibration our framework enables, applied to an object -- a mathematical model -- whose ethical expression is limited to fidelity.} 

\textcolor{black}{Third, to draw attention to subpopulations defined by the interactions of many or unfamiliar identifying characteristics is not to neglect the importance of familiar single characteristics taken in isolation, still less those already identified as posing a risk to equity. Indeed, equity with respect to a single characteristic may be concealed by disordinal interactions. For example, if compared with men of any age a model both over-performs in young women and under-performs in elderly women, calibration against sex alone will not reveal any difference between men and women at all. Even those interested in a particular characteristic cannot afford to ignore its interactions with others.}

\textcolor{black}{Fourth, since the objective is to evaluate the epistemic equity of a model with respect to a description of the population that best exposes it to scrutiny, the choice of modelled characteristics will typically be constrained only by feasibility. Prejudice for or against any given characteristic would undermine the very notion of equity we are seeking to promote. Equally, inequities embedded in the source characteristics themselves may propagate to their representation: a problem mitigated by calibrating with multiple different representations, either constructed from different sets of characteristics, or drawn from different levels of a hierarchical decomposition of a single set, yielding a multi-scale perspective on equity.}       

Fifth, although contemporary discourse on model equity is tightly focused on complex models based on machine learning \cite{bellamy2018ai,bird2020fairlearn,saleiro2018aequitas}, the question of epistemic equity arises with any model architecture -- simple, complex, transparent or opaque -- indeed with intuitive decision-making too. The simple models of traditional evidence-based medicine, crafted in ignorance of population heterogeneity, are not more equitable but merely blind to the violations of equity they may commit. It precisely because complex models seek to ground beliefs about individuals in a local feature space that variations over the population become both surveyable and remediable. But it is also true that greater model flexibility can be associated with greater model fragility, potentially amplifying disparities through subpopulation-specific catastrophic failure\cite{sagawa2019distributionally,tatman2017gender,googlemistag,amazonrecruitment,propublicarecidivism}. In any event, all models -- and intuitive decision-making -- should be evaluated, for no practice is immune to inequity.            

Sixth, the causes of epistemic inequity will vary in their susceptibility to remediation \cite{hellstrom2020bias,porta2014dictionary,olteanu2019social,torralba2011unbiased,sap2019risk,james2013introduction}. The irreducible, random component of variation may be greater in one subpopulation than another through constitutional differences in the underlying biological processes nothing could possibly equalise, such as those associated with ageing. There may be variations in data representation, class balance, labelling, and noise that are outside the power of healthcare systems to address, imposing practical, circumstantial limits on equity. But that a representation model can identify a distinct subpopulation shows its features are learnable, and that underperformance is therefore potentially addressable, even if it may not be found to be so on subsequent examination. Our framework neither assumes nor accepts differential limits on epistemic quality: its task is to focus attention on where they most need to be examined.

\textcolor{black}{Seventh, none of the components of the framework -- the target model, the representation model, the metrics of fidelity and equity, or the approach to remediation -- are generally prescribed. It is natural that each should be adapted to the specific task and its circumstances: a strength of the approach is its flexibility. In particular, the expressivity of the representation should be tuned to the heterogeneity of the data and its learnability under the applicable data and computational regime. Where supported by the regime, it is appropriate to use a finely granular representation; where not, a coarser representation may still capture systematic intersectional effects non-representational calibration would miss. Hierarchically organised representations provide a graceful way of manipulating representational expressivity without the need to retrain the representational model: one simply chooses the most suitable level of the descriptive hierarchy. In general, our innovation is not in the ingredients but in the way in which they are put together to provide a robust, comprehensive solution to the problem of quantifying equity in populations of the heterogeneity likely to obtain in reality.}

Eighth, although most naturally derived from the data on which the target model is trained, the representations used in calibration may originate from another source as long as the test data can be mapped onto the same representational space. The task of the representation model -- to redescribe the population in a way that makes its heterogeneity legible -- does not require the target label, and can therefore be accomplished with larger scale data from elsewhere. This mechanism can even be used in remediation to augment the target data to include patterns of variation derived from another source. For example, one might use learnt patterns of age-related changes in brain morphology to deform an independent set of brain images across a wider range of aged appearances \cite{pombo2021equitable}.   
    
\textcolor{black}{Ninth, remediation need not be confined to model retraining, but may encompass any action that improves the quality of the decision-making the target model is used to guide, including simply narrowing the applicable scope of a model. Theoretically the most potent action, though perhaps the least discussed in the literature, is acquisition of new data selected by its predicted impact on model equity \cite{chaloner1995bayesian,cohn1996active}. Just as such active learning may make the decision boundaries of a discriminative model easier to delineate, so it may ensure they are equitably configured. Indeed it would be entirely natural to add an equity constraint to an active learning or sequentially optimised experimental design modelling framework. In general, it should be part of the objective of remediation to attain improvements in equity \textit{without} impact on other subpopulations or on the population as a whole \cite{pombo2021equitable}. Only remediation methods that add information, through additional data or more accurate prior beliefs, could plausibly combine joint improvements in equity and overall fidelity. Since knowledge in healthcare is not a fixed quantity, to be divided more or less evenly across the population, redistributive approaches are less appropriate here than in other domains of activity \cite{berk2021fairness,hellstrom2020bias,bellamy2018ai,bird2020fairlearn,saleiro2018aequitas}. This is not a zero-sum game.}

Finally, we should recognize the deep union between epistemic equity and the individuation of care: neither is possible without the other. Equity implies maximising our knowledge of the optimal care of a patient, identified as richly as the task demands, up to the practically achievable limit; successful individuation implies having attained that maximum. Both require the flexible, highly expressive models contemporary machine learning has only recently supplied but medicine has always needed.

In summary, our proposed framework enables the assurance of the epistemic equity of any model in healthcare -- whether simple of complex -- under ethically the most general notion of identity: one defined not merely by demographics but by \textit{any} set of characteristics that define a distinct group, alone or in interaction. Our approach places ethical model calibration on a robust conceptual and algorithmic footing, advancing the application of quantitative ethics to medicine, and promoting equitable clinical care at the highest level: the knowledge it rests on.  

\section*{Methods}

\subsection*{Dataset}
The dataset for this study was drawn from the UK Biobank. The UK Biobank was established as a major prospective study with significant involvement from the UK Medical Research Council and the Wellcome Trust \cite{sudlow2015uk}, and has become an important open-access resource for medical researchers across the UK and worldwide. The subset of data contained $150,000$ records in the training set and $50,000$ records in the validation set, each record representing a distinct individual. The variables in the dataset were as follows (UK Biobank field number in brackets): Demographic: Sex (31), Age (33), Smoking (20116), Ethnicity (21000), Townsend Deprivation Index (189); Investigatory: Haemoglobin (30020), Glycated Haemoglobin (30750), Body Mass Index (21001), Weight (21002), Body Fat $\%$ (23099); Medical Diagnoses: Diabetes (2443), High blood pressure (6150),
Heart Attack / Angina / Stroke (6150), Blood Clot / Emphysema / Lung Clot, Asthma, Hayfever/Rhinitis/Eczema (6152), Other Serious Condition (2473).

The condition of interest in this study was Type II diabetes: globally a leading, and increasing, cause of morbidity and mortality \cite{vos2020global}, predicted to become the most prevalent condition in the UK Biobank cohort \cite{sudlow2015uk}. Diabetes was chosen in order to test the framework with a realistic medical problem, and a selection of variables of ethical interest. It was expected that the diagnosis of diabetes, and by association raised levels of HbA1c, would be predictable from the data. This was known to be plausible due to existing work\cite{dolezalova2021development} on diabetes prediction using UK Biobank data, which influenced the variable selection. The dataset was used for the regression task of predicting the level of glycated haemoglobin (HbA1c) using the other variables, excluding the presence of a diabetes diagnosis.

To prepare the data, a small number of records with what appeared to be outlier values of HbA1c were removed. Some other variables of interest were dropped including Income and Forced Expiratory Volume, owing to missing data. The ethnicity codes in the data were also grouped into broader categories for ease of illustration. The diagnoses of heart attack, angina and stroke were combined into one variable due to the format of the original data and low prevalence, as was Blood Clot/Emphysema/Lung Clot and Hayfever/Rhinitis/Eczema for the same reasons. Any records with missing values were removed; the number of records above refers to complete records.

\subsection*{Regression model}

A standard feed-forward neural network model was used for the regression task. Three hidden layers of neurons were used, with the number of neurons in each layer being (16,16,8), yielding 753 parameters. The ReLU activation function was used at each hidden layer. The mean squared error loss function was used. The model was trained using gradient descent with a batch size of 10 and the Adam optimiser with a learning rate of $0.001$, as has been shown to perform well in a variety of contexts \cite{kingma2014adam}. The number of epochs was set following experimentation based on error on the validation set. Variables were normalised before modelling to have a mean of zero and standard deviation of one across participants. The Pytorch library was used.

Model performance was measured using Root Mean Square Error (RMSE) and Normalised RMSE (NRMSE), which is RMSE of the model normalised by the average Root Mean Square Error of the subpopulation. This results in a proportional measure between zero and one. The normalisation was done to allow for different variance between subpopulations. For example, patients with diabetes have a higher variance in glycated haemoglobin (HbA1c), and so a higher RMSE could be expected and is not necessarily indicative of decreased model performance. One disadvantage of (N)RMSE is that it can be affected by outliers.

Existing equity metrics such as Statistical Parity Difference or Equal Opportunity Difference were not used. These tend to be classification-oriented, limited in scope and require the definition of favourable vs unfavourable outcomes. For example, judgements around the relative consequences of a disease being under- or over-diagnosed were not addressed, only that model accuracy differs across the dataset.

\subsection*{Unsupervised models}

An autoencoder model was used to in an attempt to uncover new subpopulations in the latent space of the data. An autoencoder consists of a two-part neural network, where the first ``encoder'' network compresses the data into a low-dimensional space, and the second ``decoder'' network attempts to reconstruct the original data from the compressed representation. The autoencoder architecture consisted of two hidden layers with (8,4) neurons before a ``bottleneck'' layer of two neurons (the compressed data) and a mirror-image decoder network. The mean squared error loss was used to calculate reconstruction error. For most experiments, the compressed layer was kept at two neurons for visualisation purposes, even though it was unlikely that all the information could be retained in just two variables given the specific nature of the datasets. The other training hyperparameters were set following a similar procedure to the neural network regression model. Non-binary variables were normalised before modelling.

Given a two-dimensional latent space containing key information about relationships in the dataset, a Gaussian Mixture Model (GMM) was used to identify subpopulations within that space. The GMM is a weighted mixture of multivariate Gaussian probability distributions \cite{friedman2001elements}. It can be used to estimate complex densities due to its flexibility and ability to generalise to high dimensions. Here our interest was to segment the latent space into subpopulations, which could be interpreted as probabilistic clustering where each point is assigned to the Gaussian component with the highest probability. Although GMM is effective as a clustering method, there are limitations on the shape of clusters it can find. In addition, the number of components must be specified. To determine the optimal number of components, silhouette scores were calculated. Silhouette scores are based on a comparison of distances between points both within and between clusters \cite{ogbuabor2018clustering}. They are close to one for well-separated clusters and close to zero or negative for poorly-separated clusters.
For the latent spaces in this study, the mean silhouette scores tended to decrease as the number of Gaussian components in the GMM increased. This indicates that the clusters identified by the GMMs were not well-separated, as there was no clear distinction between them. Nevertheless, the GMM was still used to segment the latent space for analysis, based on the numerical values of the latent space, which were expected to have some degree of local correspondence. A fixed number of fifty components was used to balance between diverse subpopulation identification and having a sufficient number of data points in each subpopulation. Since neural network models such as autoencoders produce distributed or ``entangled'' representations that are subject to random variation, there is no guarantee of finding meaningful subpopulations with this method, and in practice a greater degree of disentanglement would be necessary to identify consistent and meaningful subpopulations. Note this is just one possible method of segmenting the latent space: it was used here merely to exemplify the general approach to representational ethical model calibration. 

\subsection*{Subpopulation performance}

The regression model tended to perform differently across subpopulations, deviating from the ideal of equal fidelity for all. To identify the characteristics of those that exhibited particularly poor model performance, a combination of visual analysis and permutation tests was employed. Permutation tests are non-parametric tests that explore all possible random orderings of a variable in a dataset and thus produce a p-value of a variable's mean within a subpopulation \cite{efron1994introduction}. This provides an objective way to determine which variables are important in defining that subpopulation. It does not, however, give specific details and the overall results require domain knowledge to interpret. For computational reasons, it was unfeasible to calculate all possible permutations, and so the standard approximation of 1,000 rounds was used. Due to the number of tests (testing each of $25$ variables across $5$ groups), the Benjamini-Hochberg procedure \cite{benjamini1995controlling} was used to control the False Discovery Rate at $\alpha=0.05$.

To quantify the overall level of inequity in a population of model predictions, the Gini coefficient was calculated. This economic metric was originally used to index the dispersion of income differences in a population \cite{chen1982gini}. Its calculation is based on the relative mean absolute difference in incomes across the population. A Gini coefficient of zero indicates equality of income and a Gini coefficient of one indicates maximum inequality. If the distribution is random, the Gini coefficient has an expected value in the region of $0.33$. In our case, ``income'' becomes model fidelity as captured by NRMSE. The Gini coefficient is a useful summary statistic, particularly when examining the impact of a remediation algorithm on a population of model results. Note other indices of equity may be used here, dependent on the measure of fidelity most appropriate to the specific task: our use of Gini is intended to be illustrative.

\subsection*{Remediation}
\textcolor{black}{Once model inequity was identified, the chosen remediation approach was to oversample the underperforming group, before retraining the model on the rebalanced dataset. This is referred to as rebalancing. There is no consensus on the best method for remediation \cite{caton2020fairness}: rebalancing is used here merely to illustrate remediation in the context of the broader calibration framework. The underperforming group was defined to be data points in subpopulations with a performance metric below the median performance for all. This reduced the emphasis on particular subpopulations, which were automatically resampled in proportion to their size. This method was selected after previous experiments showed that rebalancing based on individual groups could push already-marginalised groups to have even worse results. Hence the dataset was split into an ``under-served'' (under-performing) group and a ``base'' (remainder) group. The oversampling multiplier was set to optimise model performance following experimentation (Figure \ref{fig:diab_hba1c_oversamp}). The experiment was repeated 10 times and the variance in results is shown in Figure \ref{fig:diab_hba1c_results} and Tables \ref{tab:diab_hba1c_results} and \ref{tab:diab_hba1c_gini}.}

\section*{Data availability}

The UK Biobank resource is available to researchers for health-related research in the public interest. All researchers who wish to access the research resource must register with UK Biobank at https://www.ukbiobank.ac.uk 

\section*{Code availability}

The code used in this study is available from the corresponding authors on request by email. 

\bibliography{sample}



\section*{Acknowledgements}

The authors would like to thank Dr Matthew Caldwell for his support and assistance, and Dr Angela Aristidou for critical comments and suggestions. This work is funded by Wellcome and the UCLH NIHR Biomedical Research Centre.

\section*{Author contributions}

R.C. and P.N. jointly developed the approach and wrote the manuscript, based on ideas formulated in discussion with I.S., J.K.R., D.H., A.N., D.B., D. F-R., and G.R. R.C. performed the experiments. All authors contributed to revising the work critically for important intellectual content, approved the completed version, and are accountability for all aspects of the work.

\section*{Competing interests}

The authors declare no competing interests.

\section*{Additional information}

\textbf{Correspondence} and requests for material should be addressed to Robert Carruthers or Parashkev Nachev.

\end{document}